\documentclass[runningheads]{llncs}

\usepackage{eccv}

\usepackage{eccvabbrv}

\usepackage{graphicx}
\usepackage{booktabs}

\usepackage{multirow}
\usepackage{bbm}
\usepackage{wrapfig}

\usepackage[accsupp]{axessibility}  

\usepackage[pagebackref,breaklinks,colorlinks]{hyperref}

\usepackage{orcidlink}

\begin{document}


\title{Alternate Diverse Teaching for Semi-supervised Medical Image Segmentation}

\titlerunning{Alternate Diverse Teaching for SSMIS}

\author{Zhen Zhao\inst{1}\orcidlink{0000-0002-0796-4078} \and Zicheng Wang\inst{3}\orcidlink{0000-0001-8351-0329} \and Longyue Wang\inst{2}\thanks{Corresponding author.}\orcidlink{0000-0002-9062-6183} \\
 Dian Yu\inst{2}\orcidlink{0000-0002-8583-8931} \and Yixuan Yuan\inst{4}\orcidlink{0000-0002-0853-6948} \and Luping Zhou\inst{3}\orcidlink{0000-0001-8762-2424}
 }

\authorrunning{Z.~Zhao et al.}

\institute{$^1$ Shanghai AI Lab \quad $^2$ Tencent AI Lab \\
$^3$ University of Sydney \quad $^4$  The Chinese University of Hong Kong \\
\url{https://github.com/zhenzhao/AD-MT}
}

\maketitle

\begin{abstract}
Semi-supervised medical image segmentation has shown promise in training models with limited labeled data. 
  However, current dominant teacher-student based approaches can suffer from the confirmation bias. To address this challenge, we propose AD-MT, an alternate diverse teaching approach in a teacher-student framework. It involves a single student model and two non-trainable teacher models that are momentum-updated periodically and randomly in an alternate fashion. 
  To mitigate the confirmation bias via the diverse supervision, the core of AD-MT lies in two proposed modules: the Random Periodic Alternate (RPA) Updating Module and the Conflict-Combating Module (CCM).
  The RPA schedules an alternating diverse updating process with complementary unlabeled data batches, distinct data augmentation, and random switching periods to encourage diverse reasoning from different teaching perspectives.
  The CCM employs an entropy-based ensembling strategy to encourage the model to learn from both the consistent and conflicting predictions between the teachers. Experimental results demonstrate the effectiveness and superiority of AD-MT on  the 2D and 3D medical segmentation benchmarks across various semi-supervised settings.
  \keywords{Semi-supervised Learning \and Medical Image Segmentation \and Alternate Diverse Teaching \and Random Periodic Alternate}
\end{abstract}

\section{Introduction}
\label{sec:intro}

\begin{figure*}
\centering
\begin{subfigure}{0.182\linewidth}
    \centering
    \includegraphics[width=0.999\linewidth]{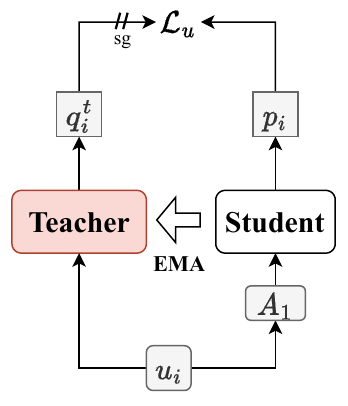}
    \caption{mean-teacher}
    \label{fig:abls:a}
\end{subfigure}
\hfill
\begin{subfigure}{0.275\linewidth}
    \centering
    \includegraphics[width=0.75\linewidth]{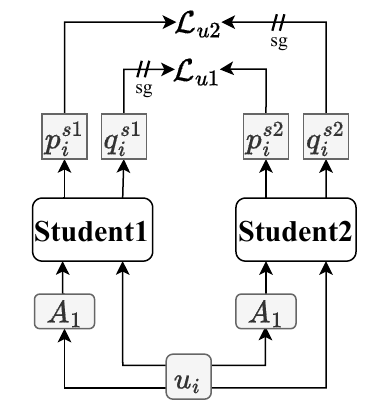}
    \caption{two-student co-training}
    \label{fig:abls:b}
\end{subfigure}
\hfill
\begin{subfigure}{0.275\linewidth}
    \centering
    \includegraphics[width=0.999\linewidth]{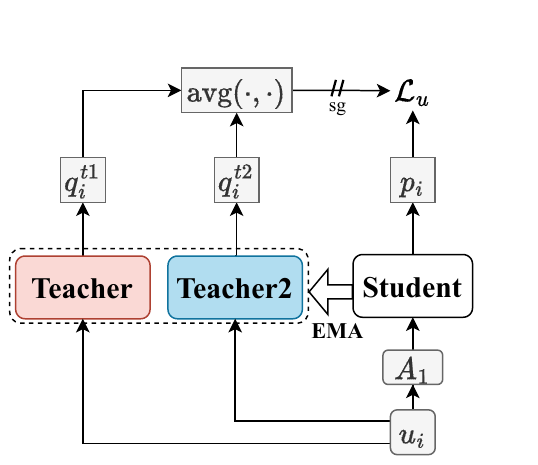}
    \caption{two-teacher ensembling}
    \label{fig:abls:c}
\end{subfigure}
\hfill
\begin{subfigure}{0.242\linewidth}
    \centering
    \includegraphics[width=0.999\linewidth]{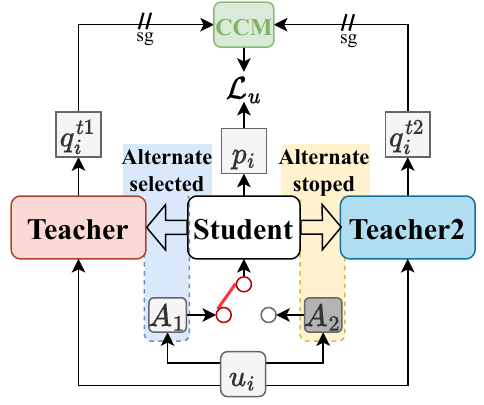}
    \caption{AD-MT}
    \label{fig:abls:d}
\end{subfigure}
\caption{Frameworks of different SSMIS methods. Crucial distinctions arise from how the unlabeled data  is leveraged. a) the plain teacher-student framework, b) a two-student co-training paradigm that enforces mutual learning,
c) a two-teacher ensemble framework where two differently initialized and updated teacher models supervise the training of the student model in a average manner, d) our proposed alternate diverse mean-teacher (\textbf{AD-MT}) framework.
Our framework involves two teacher models that are updated periodically and randomly, using complementary unlabeled data batches, distinct data augmentation strategies, and  randomized switchable periods, to enlarge their disagreement.
Additionally, our Conflict-Combating Module (CCM) encourages the student model to learn from the conflict predictions of the teacher models rather than dropping conflicts directly. ``sg" denotes ``stop gradient".}
\label{fig:abls}
\end{figure*}

Medical image segmentation is a pressing task in computer-aided diagnosis, as it plays a crucial role in medical image reasoning~\cite{valanarasu2022unext, cao2022swin, yan2022after}. Current methods for medical image segmentation heavily rely on deep neural networks, which necessitates a substantial amount of annotated data for training. However, annotating pixel-wise medical images is challenging and time-consuming, often requiring expert annotators~\cite{xiang2022fussnet, you2023action}. In response to this challenge, several studies have focused on the development of semi-supervised medical image segmentation (SSMIS) techniques, aiming to train models using a limited amount of labeled data and a larger amount of unlabeled data~\cite{luo2021dtc, wu2022ssnet, bai2023bcp}. It is evident that the key lies in effectively leveraging the unlabeled data to assist the labeled data for model training~\cite{sss22ELN, sss22PSMT, li2023cfcg}.

Recent SSMIS studies are dominated by consistency regularization (CR) based approaches, which encourage the model to generate consistent predictions from disagreements on the same unlabeled input~\cite{sss22uamt, wu2021mcnet, li2020sassnet}. Various works adopt a teacher-student framework with weak and strong data augmentation strategies and encourage the exponential moving average (EMA) teacher model to provide supervision for unlabeled training. However, as discussed in many semi-supervised studies~\cite{luo2022semi,bai2023bcp,wang2023conflict}, a single model can inevitably produce noisy and even wrong pseudo-labels, resulting in the model suffering from the so-called confirmation bias issue~\cite{bias2019}.

In the literature, early studies like DTC~\cite{luo2021dtc}, SASS-Net~\cite{li2020sassnet}, and SS-Net~\cite{wu2022ssnet}, tend to introduce extra training constraints to tackle the confirmation bias in an indirect manner. Differently, recent works turn to increasing diverse supervision signals to alleviate the bias directly. Studies along this line can be divided into two categories, \ie, the multi-student co-training method~\cite{wu2021mcnet, luo2022semi, sss21cps} and the multi-teacher ensembling method~\cite{sss22PSMT}. 
The co-training framework aims at introducing another diverse student model and encouraging both models to supervise each other mutually, as shown in~\Cref{fig:abls:b}. 
However, in addition to introducing extra training efforts and losing the EMA stability, these methods using the same network structure cannot produce a sufficient discrepancy between co-training models by only using different initialization and learning rates.
On the other hand, the multi-teacher ensembling methods encourage the student model to update different teacher models iteratively, avoiding extra training costs, as shown in~\Cref{fig:abls:c}. 
However, a key issue in such methods is the teacher model updating strategy, which has not been carefully designed to generate diverse supervision.
Besides, existing ensembling methods typically adopt an average strategy and train the model only from their consistent predictions. Few studies have explored the potential benefits of learning from the conflicts.

\begin{wrapfigure}{r}{7cm}
\centering
\includegraphics[width=0.99\linewidth]{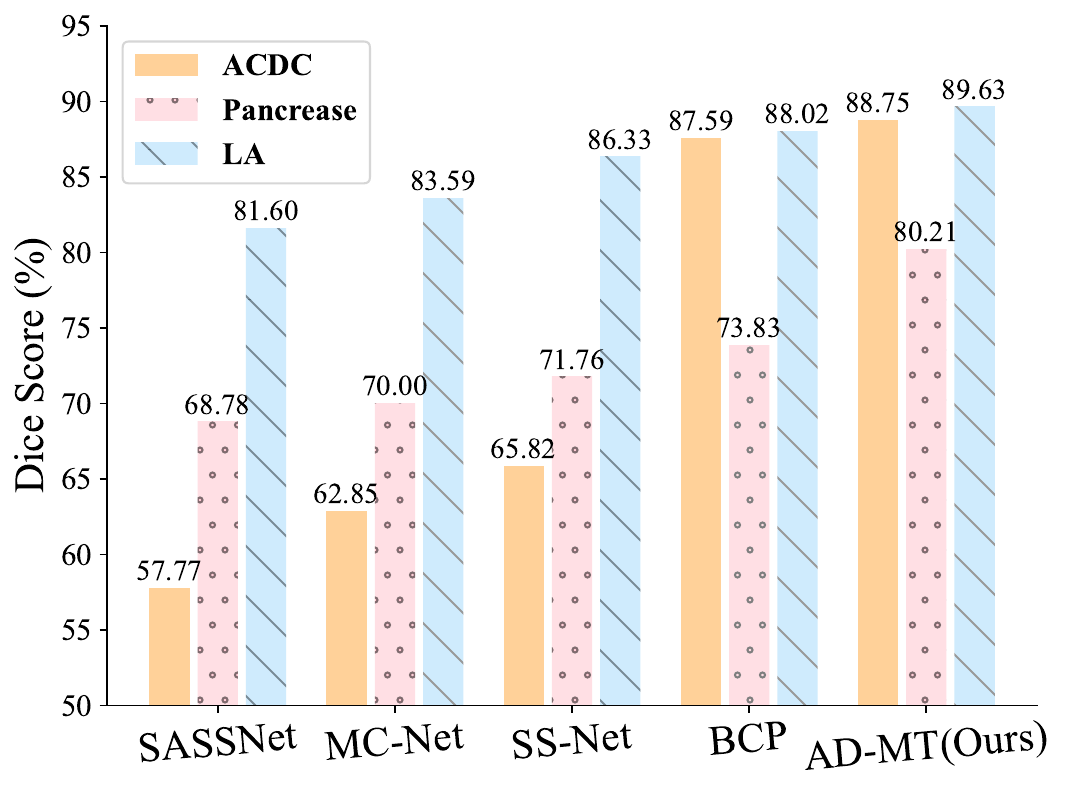}
\caption{We compare our proposed AD-MT with recent SSIMS methods in terms of the Dice score on 2D ACDC, 3D LA and Pancreas datasets with 3, 4, and 6 labeled instances, respectively. Our end-to-end AD-MT can consistently outperform the current state-of-the-art BCP (\textit{which requires an additional pre-training stage}).}
\label{fig:intro}
\end{wrapfigure}

Facing these issues, in this paper, we propose a novel alternate diverse teaching approach in a teacher-student framework, dubbed AD-MT, for SSMIS. As shown in \Cref{fig:abls:d}, AD-MT involves a single trainable student model and two non-trainable teacher models that are updated directly by the EMA of the student weights. 
In specific, two teachers are updated periodically and randomly in an alternate fashion. To encourage diverse reasoning from different teaching perspectives, AD-MT enlarges the discrepancy by using complementary data batches, distinct data augmentation strategies, and randomized switchable periods. This alternating diverse updating process is scheduled by our proposed Random Periodic Alternate Updating (RPA) Module.
Furthermore, instead of discarding conflicting predictions of the teacher models, we design an entropy-based Conflict-Combating Module (CCM) that separates the consistent and conflicting predictions and also encourages the model to learn from the disagreements between the teachers. Thanks to our proposed two modules, AD-MT, as shown in \Cref{fig:intro}, can consistently outperform current state-of-the-art (SOTA) BCP by a large margin, achieving a 6.38\% Dice improvement on the Pancrease with 10\% labeled data. Our contributions are summarized as follows,
\begin{itemize}
    \item We propose AD-MT, an alternate diverse teacher-student approach for SSMIS, which enforces diverse teaching models to mitigate the impact of confirmation bias.
    \item We design a novel Random Periodic Alternate updating module to enlarge the diversity of the two teacher models and a Conflict-combating module to learn from both consistent and conflicting predictions of the two teachers.
    \item Without introducing additional constraints and extra training costs, AD-MT achieves the new SOTA performance on the 2D and 3D SSMIS benchmarks.
\end{itemize}

\section{Related work}
\label{sec:rwork}

As highlighted by various works~\cite{semifix,zhao2023entropy,ssl22dcssl,ssl22lassl,gui2022improving,duan2022mutexmatch}, the effective utilization of unlabeled data plays a crucial role in tackling the semi-supervised problem. Among the existing studies, consistency regularization (CR) based approaches have emerged as the dominant direction in SSMIS. These methods aim to produce disagreements on the same unlabeled inputs, thereby training the model to generate consistent predictions~\cite{yang2023unimatch, sss22PSMT, sss22st++,sss21simple,sss22reco,zhao2023augmentation}. 
Previous methods along this line have focused on introducing perturbations and generating pseudo-labels using stable predictions as supervision for unstable ones~\cite{ssl18vat,sss22st++,zhao2023instance}.
Mean-teacher~\cite{semimt}, as shown in~\Cref{fig:abls:a},  is a widely adopted semi-supervised learning framework in SSMIS studies. Latter studies explored the significance of weak and strong augmentation strategies to produce sufficient prediction disagreement~\cite{semiuda,semifix,zhao2023augmentation,bai2023bcp}. However, the supervision signal derived from the predictions of the unlabeled data is inherently noisy, which can lead to an issue known as confirmation bias~\cite{bias2019}. 
The bias issue can negatively impact the training stability and hinder the model's recognition ability. On top of Mean-teacher, UAMT~\cite{sss22uamt} proposed an uncertainty-aware training scheme to alleviate the bias.
SASS-Net~\cite{li2020sassnet} posed a geometric shape constraint upon the segmentation outputs, and SSNet~\cite{wu2022ssnet} designed additional contrastive losses to enhance the model's discriminative ability, which both tended to improve the SSMIS in an indirect manner.

Differently, recent works propose increasing the supervision signals to mitigate the potential bias and improve the training process directly. These approaches can be broadly categorized into two main categories: the multi-student co-training methods and the multi-teacher ensembling methods. As illustrated in~\Cref{fig:abls:b}, two-student co-training methods~\cite{wu2021mcnet, luo2022semi, wang2023conflict} tackle the noisy supervision problem by introducing an additional branch to the learning framework.  
The incorporation of an extra branch involves another student model, which can mutually supervise each other. Such an approach encourages diverse reasoning from different perspectives, thereby mitigating the impact of confirmation bias. 
Existing studies along this line typically adopt the different model initialization and learning rates to maintain the discrepancy between student models~\cite{other20dual,sss21cps}. Differently, our proposed AD-MT enlarges the discrepancy by using complementary sets of unlabeled data batches, distinct data augmentation strategies, and randomized switchable periods while reserving the benefits of exponential moving averaging teacher models. Besides, these methods also come at the cost of increased training costs, as they introduce additional training parameters.

On the other hand, some works adopt the multi-teacher ensembling framework~\cite{sss22PSMT, na2023switching}, as shown in~\Cref{fig:abls:c}. 
Along this line, the student model iteratively updates multiple teacher models with different updating strategies, leveraging their differing perspectives. 
By utilizing multi-teacher models, these methods ensure supervision diversity without introducing additional training parameters. 
However, it is worth noting that existing methods have not carefully considered the updating strategy to enforce the teacher models to be sufficiently different. For example, some works only utilize different initialization or updating at different epochs to encourage teacher differences. It is important to develop more effective and robust updating strategies that can maintain the diversity of the teacher models while ensuring training stability.
More importantly, it is also crucial to address the issue of conflicting supervision when tackling the ensembled predictions. 
Conflicting supervision naturally occurs when different sources of supervision provide contradictory guidance to the student model, potentially leading to training instability and suboptimal segmentation results. 
Despite the significance, most of the existing methods tend to drop these conflicts but have not explicitly tackled the issue of conflicting supervision.

\begin{figure*}[t]
    \centering
    \includegraphics[width=0.92\linewidth]{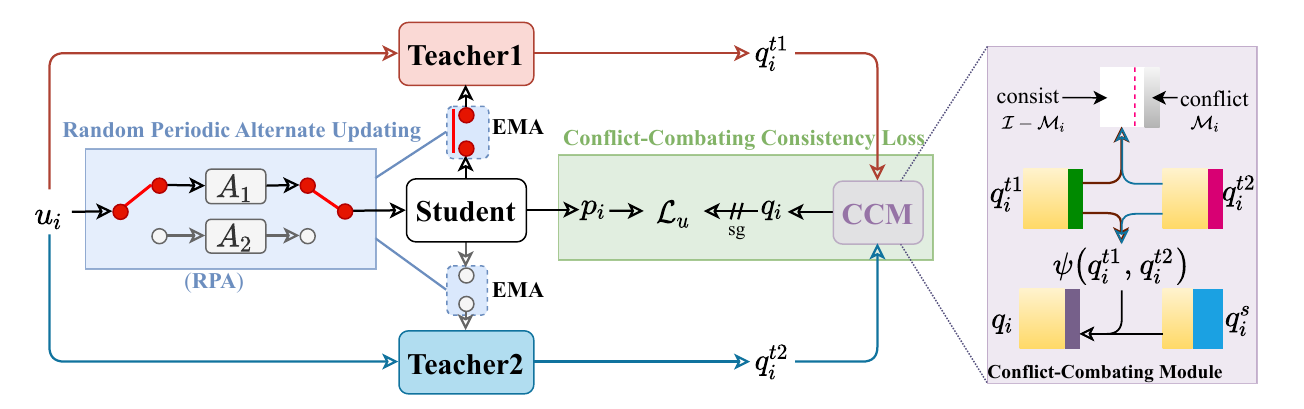}
    \caption{The diagram of our proposed AD-MT. Our method consists of two main modules: the Random Periodic Alternate Updating Module (RPA) and the Conflict-combating Module (CCM). Specifically, two teacher models T1 and T2 are updated in turn periodically and randomly. At each iteration, only one certain teacher model $\mbox{T}_m$, ($m=1,2$) will be updated, using complementary unlabeled data batches and different strong data augmentation strategies $A_m$ accordingly. Furthermore, the switchable period of two teachers is randomly generated by the RPA module, aiming to increase the disagreement between the two teacher models. Meanwhile, the CCM module separates the consistent and conflicting  predictions of two teacher models, and encourage the model to learning from instead of dropping the conflicts. $q_i^{s}, q_i^{t_1}, q_i^{t_2}$ represent the generated pseudo-labels from the student and two teachers models, respectively.  
    }
    \label{fig:diagram}
\end{figure*}

\section{Method}
\label{sec:method}

In this section, we provide an overview of our AD-MT, followed by a detailed description of its two main components: the Random Periodic Alternate Updating Module (RPA) and the Conflict-Combating Module (CCM).

\subsection{Overview}

In the context of semi-supervised medical image segmentation (SSMIS), the available data comprises both labeled samples $\mathcal{X}$ and unlabeled samples $\mathcal{U}$, with the number of labeled samples typically being much smaller than that of unlabeled ones (\ie, $|X| \ll |U|$). During the training process, 
given a batch of labeled samples $\mathcal{B}_x= \{(x_i, y_i)\}_{i=1}^{B}$ and a batch of unlabeled samples $\mathcal{B}_u=\{u_i\}_{i=1}^{\mu B}$, SSMIS methods aim to obtain a good segmentation model by leveraging both labeled and unlabeled data effectively. 
Different from widely-adopted co-training methods, our proposed method, as illustrated in \Cref{fig:diagram}, utilizes a single student model, parameterized by $\theta_s$, that receives model back-propagating gradient. Besides, two auxiliary teacher models, parameterized by $\theta_{t_1}$ and $\theta_{t_2}$, are alternatively updated by the exponential moving average (EMA) of the student model weights. Similar to the plain teacher-student methods~\cite{semimt}, the student model can be directly trained on the labeled data via a standard supervised loss $\mathcal{L}_x$, 
\begin{align}
    \mathcal{L}_x = \frac{1}{|\mathcal{B}_x|} \sum_{i=1}^{B} \frac{1}{H\times W}\sum_{j=1}^{H\times W} \ell(\hat{y}_i(j), y_i(j)),
\end{align}
where $\hat{y}_i$ denotes the student model's prediction on the $i$-th labeled data $x_i$, \textit{i.e.,} $\hat{y}_i\!=\!f(x_i; \theta_s)$, and $j$ represents the $j$-th pixel on the image or the corresponding segmentation mask with a resolution of $H \times W$. Following~\cite{luo2021dtc,bai2023bcp}, $\ell$ represents the loss function, calculated by an average of the dice and cross-entropy loss. 

Though Teacher1 (T1) and Teacher2 (T2) are updated in turn, both models are involved in generating pseudo-labels for unlabeled data at each iteration simultaneously. Using $a(\cdot)$ denote the weak augmentations, which include random cropping and flipping operations~\cite{sss22uamt,wu2022ssnet}, we can obtain pseudo-labels for each unlabeled instance $u_i$, \ie, $q_i^{t_1} = f(a(u_i); \theta_{t_1})$, $q_i^{t_2} = f(a(u_i); \theta_{t_2})$, and $q_i^s = f(a(u_i); \theta_s)$ from two teacher models as well as the student model, respectively. Our proposed AD-MT can then exploit all this information and obtain an ultimate pseudo-label $q_i$,
\begin{align}
    q_i = \phi(q_i^{t_1}, q_i^{t_2}, q_i^s, \tau),
\end{align}
where $\phi(\cdot)$ represents a function of our proposed \textbf{Conflict-Combating Module}, and $\tau$ denotes a pre-defined high-confidence threshold. Subsequently, our method employs a conflict-combating consistency loss, $\mathcal{L}_u$, on unlabeled data, 
\begin{align}
    \mathcal{L}_{u} = \frac{1}{|\mathcal{B}_u|} \sum_{i=1}^{\mu B} \frac{1}{H\times W}\sum_{j=1}^{H\times W} \ell(p_i(j), q_i(j)), 
\end{align}
where $p_i\!=\!f(A_m(u_i);\theta_s)$ is the student model's prediction on strongly-augmented unlabeled data $A_m(u_i)$. $A_m\!\in\!\{A_1, A_2\}, m\!=\!1,2.$ represents two different strong data augmentation strategies, corresponding to the alternate turn of T1 and T2, respectively. The whole updating strategy is performed by our \textbf{Random Periodic Alternate Updating Module}. In summary, the total training loss is,
\begin{align}
    \mathcal{L} = \mathcal{L}_x + \lambda_t \mathcal{L}_u,
\end{align}
where $\lambda_t$ denotes an iteration-dependent function to adjust the importance of consistency loss $\mathcal{L}_u$. Apart from the overall straightforward structure, as shown in \Cref{fig:diagram}, the core of AD-MT lies in two aspects. First, two teachers are updated periodically and randomly in an alternate fashion. During the training procedure, we utilize complementary sets of unlabeled data, different data augmentation strategies, and randomized switching periods to ensure that the two teacher models are updated in a distinct manner, complementing each other throughout the process. Second, instead of discarding conflicting predictions of the two teacher models, we design a Conflict-Combating strategy that encourages the model to learn from the disagreements between the teachers. This strategy proves highly beneficial in improving segmentation performance at the latter stages of training. These two aspects correspond to our proposed two novel components, the RPA and the CCM, which are detailed in the following two sections.

\subsection{Random Periodic Alternate Updating Module}
\label{sec:method:alt}
Maintaining two different teacher models can benefit the pseudo-labeling and further reduce the confirmation bias in semi-supervised learning. However, to maximize the benefits of multiple models, it is essential to ensure that they are as diverse as possible. To this end, in AD-MT, we propose the Random Periodic Alternate Updating Module, a novel approach to updating two teacher models in a way that maximizes their diversity with little additional training efforts. In specific, the RPA module involves the following strategies.
\begin{itemize}
    \item \textbf{Alternate Updating.} 
    At each iteration, only one of the two teacher models is updated. Consequently, throughout a complete training epoch, unique and complementary batches of unlabeled data are utilized to refine the two teacher models. This strategy ensures that the updates to the teacher models are distinct, allowing them to complement each other's learning across the training process.
    We denote the alternate updating period (in the unit of iterations) of each teacher model by $\mathcal{T}_m$, where $m\in\{1,2\}$. 
    \item \textbf{Distinct augmentation strategies.} To further increase the diversity between the two teacher models, we also employ distinct augmentation strategies in addition to using different data batches. Specifically, we apply the color-jitting~\cite{augs20randaugment} operation for the turn of T1 while the copy-paste~\cite{augs21copy} augmentation for the turn of T2.
    \item \textbf{Randomized switching periods.}  Instead of adopting a fixed switchable period rigidly for each teacher model, we consider increasing the randomness of the switchable pattern. Given a pre-defined maximum value of the period, denoted by $\mathcal{T}_{max}$, the alternating period  for each teacher is randomly generated when switching occurs, \ie,
    \begin{align}
        \mathcal{T}_m &\leftarrow \mbox{random.randomint}(0, \mathcal{T}_{max}), \quad m \in \{1,2\}.
    \end{align}
\end{itemize}
In this way, our RPA module cannot only increase the diversity between the two teacher models but also lower the risks of data over-perturbation discussed in ~\cite{zhao2023augmentation,sss21simple}. This is simply because these different strong augmentations are not applied simultaneously in our method.

\subsection{Conflict-combating Module}
\label{sec:method:ccm}

Having two distinct teacher models derived from the RPA, our proposed Conflict-Combating Module, abstracted as a function of $\phi(\cdot)$, is carefully designed to address the issue of conflicting predictions between the two teacher models. Instead of discarding these conflicts directly, the CCM module encourages the model to further learn from the disagreements with the help of the student model.

Specifically, the CCM module first separates the consistent and conflicting predictions of the two teacher models. On the one hand, it applies the entropy-based teacher ensemble to obtain an ensembled prediction, $\psi_i$, on the unlabeled instance $u_i$, which is directly used for the consistent supervision. Given the prediction entropy (denoted by $H(\cdot)$) of two teacher models, 
\begin{align}
    H_{t_1} &= H(q_{i}^{t_1}) = -\sum_{i=1}^C q_{i}^{t_1} \log_2 q_{i}^{t_1}, \\
    H_{t_2} &= H(q_{i}^{t_2}) =-\sum_{i=1}^C q_{i}^{t_2} \log_2 q_{i}^{t_2},
\end{align}
we can obtain the entropy-based ensembled prediction,
\begin{align}
\psi_i &= \psi(q_{i}^{t_1}, q_{i}^{t_2}) = \frac{w_1 q_{i}^{t_1} + w_2 q_{i}^{t_2}}{w_1 + w_2},
\end{align}
with $w_1 = e^{-H_{t_1}}, w_2 = e^{-H_{t_2}}$. On the other hand, to account for the increasing improvement of the student model, we compare the entropy of the student's prediction with the entropy of the ensembled prediction. We then use the lower-entropy one as the final supervision for the conflicting prediction. This strategy ensures that the conflicting supervision benefits from the strengths of both the teacher models and student model. In summary, the final prediction $q_i$ is,
\begin{align}
    q_i(j)\!=\!
\begin{cases}
\mathbbm{1}(\max(q_i^s(j))\!\geq\!\tau) q_i^s(j),  &q_i^{t_1} \neq  q_i^{t_2} \,\&\, H_{\psi_i(j)} > H_{q_i^s(j)} \\
\mathbbm{1}(\max(\psi_i(j))\!\geq\!\tau)\psi_i(j), &\mbox{otherwise}
\end{cases}
\end{align}
where $\mathbbm{1}(\cdot)$ only selects the high-confidence predictions for the unlabeled supervision. Additionally, we examine more ensembling strategies in the experiment section. As the training process progresses, the student model learns from both diverse teacher models efficiently and effectively. In the inference phase, the student model is used as the final segmentation model, providing accurate and reliable segmentation results on new medical images.


{ 

\begin{table*}[t]
\centering 
\caption{Performance comparison with the SOTA methods on the \textbf{LA}, with 5\% and 10\% labeled data. $^\dag$ denotes that BCP~\cite{bai2023bcp} requires an additional pre-training stage before the semi-supervised training. The best is highlighted in \textbf{Bold}.} 
\label{tab:exp:LA}
\resizebox{0.95\textwidth}{!}{
\begin{tabular}{c|cccc|cccc}
\toprule
\multirow{2}{*}{Method} & \multicolumn{4}{c|}{Left atrium (5\% / 4 labeled data)}                                       & \multicolumn{4}{c}{Left atrium (10\% / 8 labeled data)}                                         \\ \cline{2-9} 
&Dice~$\uparrow$ & Jaccard~$\uparrow$ & 95HD~$\downarrow$& ASD~$\downarrow$ &Dice~$\uparrow$ & Jaccard~$\uparrow$ & 95HD~$\downarrow$& ASD~$\downarrow$ \\ \hline

VNet (SupOnly) & 52.55 &39.60 &47.05 &9.87&82.74 &71.72 &13.35 &3.26 \\ \hline
UA-MT\cite{sss22uamt}~(MICCAI'19) &82.26 &70.98 &13.71 &3.82 &86.28 &76.11 &18.71 &4.63 \\ 
    
SASSNet\cite{li2020sassnet}~(MICCAI'20)  &81.60 &69.63 &16.16 &3.58 &85.22 &75.09 &11.18 &2.89  \\ 
    
DTC\cite{luo2021dtc}~(AAAI'21)&81.25 &69.33 &14.90 &3.99 &87.51 &78.17 &8.23 &2.36  \\

URPC\cite{luo2021urpc}~(MedIA'22) &82.48 &71.35 &14.65 &3.65 &85.01 &74.36 &15.37 &3.96\\

SS-Net\cite{wu2022ssnet}~(MICCAI'22)&86.33 &76.15 &9.97 &2.31 &88.55 &79.62 &7.49 &1.90\\
    
MC-Net+\cite{wu2021mcnet}~(MedIA'22) &83.59 & 72.36 &14.07 &2.70& 88.96     & 80.25   &  7.93    &1.86       \\
PS-MT\cite{sss22PSMT}~(CVPR'22)& 88.49     & 79.13       & 8.12    &  2.78    & 89.72    & 81.48  & 6.94   &  1.92   \\ 
    
MCF\cite{wang2023mcf}~(CVPR'23)& -     & -       & -    &  -    & 88.71    & 80.41  & 6.32    &  1.90   \\ 
BCP\cite{bai2023bcp}$^\dag$~(CVPR'23) &88.02 & 78.72 &7.90 &2.15 & 89.62     & 81.31   &  6.81    &1.76 \\ \hline
    
\textbf{AD-MT} (Ours)& \textbf{89.63} &  \textbf{81.28} & \textbf{6.56} & \textbf{1.85} & \textbf{90.55}     &  \textbf{82.79}   &   \textbf{5.81}    & \textbf{1.70} \\ 
\bottomrule
\end{tabular}
}
\end{table*}
}

{ 

\begin{table*}[t]
\centering 
\caption{Performance comparison with the SOTA methods on the \textbf{ACDC}, in the semi-supervised setting of 5\% and 10\% labeled data. }
\label{tab:exp:acdc}
\resizebox{0.95\textwidth}{!}{
\begin{tabular}{c|cccc|cccc}
\toprule
\multirow{2}{*}{Method} & \multicolumn{4}{c|}{ACDC (5\% / 3 labeled data)}                                       & \multicolumn{4}{c}{ACDC (10\% / 7 labeled data)}                                         \\ \cline{2-9} 
&Dice~$\uparrow$ & Jaccard~$\uparrow$ & 95HD~$\downarrow$& ASD~$\downarrow$ &Dice~$\uparrow$ & Jaccard~$\uparrow$ & 95HD~$\downarrow$& ASD~$\downarrow$ \\ \hline

U-Net (SupOnly) &47.83 &37.01 &31.16 &12.62 &79.41 &68.11 &9.35 &2.70 \\ \hline

UA-MT\cite{sss22uamt}~(MICCAI'19) &46.04 &35.97 &20.08 &7.75    &81.65 &70.64&6.88 &2.02 \\ 
    
SASSNet\cite{li2020sassnet}~(MICCAI'20)  &57.77 &46.14 &20.05 &6.06    &84.50 &74.34 &5.42 &1.86  \\ 
    
DTC\cite{luo2021dtc}~(AAAI'21)&56.90 &45.67 &23.36 &7.39   &84.29 &73.92 &12.81 &4.01  \\

CPS\cite{sss21cps}~(CVPR'21)&70.15 & 61.17 & 5.96 & 2.14   &86.91 &78.11 &5.72 &1.92\\

URPC\cite{luo2021urpc}~(MedIA'22)&55.87 &44.64 &13.60 &3.74   &83.10 &72.41 &4.84 &1.53\\

SS-Net\cite{wu2022ssnet}~(MICCAI'22)&65.82 &55.38 &6.67 & 2.28  &86.78 &77.67 &6.07 &1.40\\

MC-Net+\cite{wu2021mcnet}~(MedIA'22) &62.85 &52.29 &7.62 &2.33 & 87.10     & 78.06   &  6.68    &2.00       \\

PS-MT\cite{sss22PSMT}~(CVPR'22) &86.94 & 77.90 &4.65 &2.18  & 88.91     & 80.79     &4.96 &  1.83   \\

BCP\cite{bai2023bcp}~(CVPR'23) &87.59 & 78.67 &1.90 &0.67 & 88.84     & 80.62   &  3.98    &1.17 \\ \hline

\textbf{AD-MT} (Ours)& \textbf{88.75} & \textbf{80.41} & \textbf{1.48} & \textbf{0.50} &      \textbf{89.46} &  \textbf{81.47} & \textbf{1.51}      & \textbf{0.44} \\ 
\bottomrule
\end{tabular}
}
\end{table*}

}

\section{Experiments}
\label{sec:exps}

\subsection{Datasets and Evaluation Metrics}
Following the previous works~\cite{wu2022mcnet+,wu2022ssnet,bai2023bcp}, we adopt the widely used benchmarks, the \textbf{Pancreas-NIH} dataset~\cite{roth2015deeporgan}, the  Left Atrium (\textbf{LA}) dataset~\cite{xiong2021global}, and the Automated Cardiac Diagnosis Challeng (\textbf{ACDC}) dataset~\cite{dataacdc} to validate the effectiveness of our proposed AD-MT. The Pancreas-NIH and LA are two 3D datasets, consisting of  82 contrast-enhanced abdominal CT volumes and 100 3D gadolinium-enhanced magnetic resonance image scans, respectively. 
During the training stages, the 3D images are randomly cropped into 112x112x80 and 96x96x96 for Pancrease and LA, respectively.
The ACDC dataset is a 2D benchmark, which contains 100 cardiac MR imaging samples, which are resized into 256×256 pixels and normalized into [0, 1].

Consistent with previous studies \cite{wu2022ssnet, luo2021dtc, luo2021urpc}, we adopt the Dice Score (\%), Jaccard Score (\%), 95\% Hausdorff Distance in voxel (95HD), and Average Surface Distance in voxel (ASD) as our evaluation metrics to compare the segmentation performance under the different semi-supervised partition protocols. A higher Dice Score and Jaccard Score indicate better segmentation performance, while a lower 95HD and ASD indicate better agreement between the predicted segmentation and ground truth.


{ 
\begin{table*}[t]
\centering 
\caption{Performance comparison with SOTA methods on the \textbf{Pancreas}, in the semi-supervised setting of 10\% and 20\% labeled data. }
\label{tab:exp:pancrease}
\resizebox{0.95\textwidth}{!}{
\begin{tabular}{c|cccc|cccc}
\toprule
\multirow{2}{*}{Method} 
& \multicolumn{4}{c|}{Pancreas (10\% / 6 labeled data)}          
& \multicolumn{4}{c}{Pancreas (20\% / 12 labeled data)}      \\\cline{2-9} 
&Dice~$\uparrow$ & Jaccard~$\uparrow$ & 95HD~$\downarrow$& ASD~$\downarrow$ &Dice~$\uparrow$ & Jaccard~$\uparrow$ & 95HD~$\downarrow$& ASD~$\downarrow$ \\ \hline

VNet (SupOnly) & 55.60     & 41.74           & 45.33   & 18.63 & 72.38     & 58.26    & 19.35  & 5.89\\ \hline

UA-MT\cite{sss22uamt}~(MICCAI'19)& 66.34     & 53.21      & 17.21   & 4.57        & 76.10     & 62.62   & 10.84  & 2.43        \\
    
SASSNet\cite{li2020sassnet}~(MICCAI'20)& 68.78     & 53.86     &  19.02   &  6.26        &  77.66     & 64.08   &  10.93 & 3.05       \\ 
    
DTC\cite{luo2021dtc}~(AAAI'21)&69.21    & 54.06    &  17.21    & 5.95        & 78.27     & 64.75 &  8.36  & 2.25      \\
    
ASE-Net\cite{lei2022asenet}~(TMI'22)& 71.54     & 56.82     &  16.33    & 5.73       & 79.03     & 66.57 &  8.62  & 2.30     \\ 

SS-Net\cite{wu2022ssnet}~(MICCAI'22)& 71.76     & 57.05     & 17.56    & 5.77       & 78.98     & 66.32  & 8.86    & 2.01       \\
    
MC-Net+\cite{wu2021mcnet}~(MedIA'22)& 70.00     & 55.66     &  16.03    & 3.87     & 79.37     & 66.83 &  8.52  & 1.72       \\
PS-MT\cite{sss22PSMT}(CVPR'22)& 76.94     & 62.37    &  13.12 & 3.66   & 80.74     & 68.15  & 7.41    &  2.06    \\

MCF\cite{wang2023mcf}~(CVPR'23)& -     & -       & -    &  -    & 75.00     & 61.27  & 11.59    &  3.27    \\ 

BCP\cite{bai2023bcp}~(CVPR'23) & 73.83     & 59.24       &  12.71  & 3.72       & \textbf{82.91}     &\textbf{70.97} &  6.43   & 2.25     \\ \hline
    
\textbf{AD-MT} (Ours) & \textbf{80.21}     &    \textbf{67.51}     &   \textbf{7.18}   &   \textbf{1.66} & 82.61    &70.70 &  \textbf{4.94}   & \textbf{1.38}       \\ \bottomrule
\end{tabular}
}
\end{table*}
}

\subsection{Implementation Details}
Following other SSMIS studies~\cite{sss22uamt,wu2022mcnet+,bai2023bcp}, we adopt the U-Net~\cite{seg15unet} and V-Net~\cite{milletari2016v} as the backbones for the experiments on 2D and 3D datasets, respectively.
For the 2D ACDC dataset, we train the segmentation model with a batch size of 24 (12 labeled and 12 unlabeled instances) for 30,000 iterations.
On the LA and Pancreas datasets, we follow existing studies and adopt a batch size of 4 (2 labeled and 2 unlabeled instances) for training 15,000 iterations.
We use an SGD optimizer to train the student model with a polynomial learning-rate decay where the initial learning rate, 0.01, is multiplied by $(1 - \mbox{iter}/\mbox{max\_iter})^{0.9}$. The momentum and the weight decay are set as 0.9 and 0.0001, respectively. The two teacher models are randomly initialized and updated with an exponential parameter of 0.99.
By default, we set the maximum loss weight $\lambda_u = 2.0$, and the maximum period $\mathcal{T}_{max} = 0.5\, \mbox{epoch}$ for all runs.

\subsection{Comparison with SOTAs}
In this section, we compare our method with the most recent SSMIS methods, including UA-MT~\cite{sss22uamt}, SASSNet~\cite{li2020sassnet}, DTC~\cite{luo2021dtc}, ASE-Net~\cite{lei2022asenet}, SS-Net~\cite{wu2022ssnet}, MC-Net+~\cite{wu2022mcnet+}, MCF~\cite{wang2023mcf}, PS-MT\cite{sss22PSMT} and BCP~\cite{bai2023bcp}. Note that BCP requires an additional pre-training stage while other methods do not.

\textbf{3D LA}. 
As shown in~\Cref{tab:exp:LA}, our proposed AD-MT approach achieves the highest Dice Score and Jaccard Score on both the 5\% and 10\% labeled data settings for the LA dataset. 
With only 4 labeled instances, our approach achieves a Dice Score of 89.63\%, outperforming the previous state-of-the-art method BCP \cite{bai2023bcp} by 1.61\% in Dice Score and 2.56\% in Jaccard Score, without introducing an extra pre-training stage. 
Also note that our AD-MT outperforms MC-Net+ by over 6\% in terms of the Dice score with 4 labeled data available, despite the fact that MC-Net+ introduces far more training parameters.
Meanwhile, AD-MT also obtains the lowest 95HD and ASD on both labeled data settings, indicating that our approach produces better segmentation results that are closer to the ground truth than the other SOTA methods.
These results demonstrate the effectiveness of our proposed approach in improving the accuracy of SSMIS, even in scenarios where labeled data is scarce.

\textbf{2D ACDC}.
\Cref{tab:exp:acdc} shows the results of our AD-MT compared to the current SOTA methods on the ACDC dataset. The ACDC dataset is a challenging dataset for SSMIS due to fine-grained multiple classes and the variability in heart anatomy and pathology, making it a good benchmark for evaluating the effectiveness of our approach.
Similar to the results on the LA, AD-MT achieves the best performance on the ACDC dataset in both the 5\% and 10\% labeled data settings. In the 5\% labeled data setting, AD-MT achieves a Dice Score of 88.75\%, which is 1.16\% higher than BCP's Dice Score of 87.59\%.
A notable advantage of our proposed AD-MT approach is that it does not require any additional pre-training stage before the semi-supervised training, unlike the BCP method \cite{bai2023bcp}. This makes AD-MT a more efficient and practical approach.
AD-MT achieves a significant improvement in Dice Score compared to other end-to-end methods without pre-training, with more than a 20\% improvement observed in the 5\% labeled setting.

\textbf{3D Pancreas}. 
The results on the Pancreas-NIH dataset are reported in~\Cref{tab:exp:pancrease}. It can be seen from the table that our method demonstrates great recognition performance when the number of labeled data is small, \eg, our AD-MT surpasses the baseline method and UA-MT by a large margin of more than 20\% and 10\% in terms of the Dice score with 6 labeled data available, respectively. 
In the 10\% labeled setting, AD-MT achieves a Dice Score of 80.21\%, which is 6.38\% higher than the BCP's Dice Score of 73.83\%. 
In the 20\% labeled setting, AD-MT achieves a Dice Score of 82.61\%, which is only slightly lower than the BCP's Dice Score of 82.91\%. However, AD-MT achieves the lowest 95HD and ASD, indicating that our approach produces more dedicated segmentation results. Similar to the observation on the LA and ACDC, our AD-MT produces highly accurate segmentation on the Pancreas with limited labeled data, which is a significant advantage in real-world scenarios where labeled data is scarce and expensive to obtain.

{ 
\begin{table*}[t]
\centering
\caption{Ablation studies on different components of our proposed AD-MT, when using 3 cases as labeled data on the ACDC. Three different classes of RV, Myo, LV represent the right ventricle, myocardium, and left ventricle, respectively.}
\label{ablation:component}
\resizebox{0.95\textwidth}{!}{
\begin{tabular}{cccc|cc|cc|cc|cc}
\toprule
\multicolumn{4}{c|}{\textbf{Components}}&\multicolumn{2}{c}{\textbf{RV}}&\multicolumn{2}{c}{\textbf{Myo}}&\multicolumn{2}{c}{\textbf{LV}}&\multicolumn{2}{|c}{\textbf{Mean}} \\\cline{1-12} 
T1 & T2& RPA & CCM &
Dice(\%)&95HD&
Dice(\%)&95HD&
Dice(\%)&95HD&
Dice(\%)$\uparrow$&95HD$\downarrow$\\ 
\hline
\checkmark& & & & 85.30 & 2.18 & 84.44 &1.53 & 90.76 &  4.25 & 86.83 & 2.65\\
& \checkmark & & & 84.59 & 3.08 & 83.51 & 1.86 & 90.56 &  2.33 & 86.22 & 2.43\\
\checkmark & \checkmark& \checkmark & & 85.80 & 1.98 & 85.63 & 1.29 & 92.20 &  2.83 & 87.88 & 2.03\\
\checkmark& \checkmark&\checkmark& \checkmark& \textbf{86.63} &  \textbf{1.92} & \textbf{86.78} & \textbf{1.17} & \textbf{92.86} & \textbf{1.36} & \textbf{88.75} & \textbf{1.48}\\
\bottomrule
\end{tabular}
}
\end{table*}
}

{ 
\begin{table}[t]
    \centering
    \caption{Ablation studies on the threshold $\tau$ with 5\% labeled data. It is set as $0.95$ and $0.75$ for the 2D and 3D datasets, respectively. }
    \label{tab:abl:threshold}
    \begin{tabular}{c|cccccc}
    \toprule
    $\tau$ & 0.7 & 0.75 & 0.8 & 0.85 & 0.9 & 0.95 \\
    \midrule
    ACDC    & 86.58  & 87.90 & 87.97 & 88.07 & 88.42 & \textbf{88.75} \\
    LA    & 89.59 & \textbf{89.63} & 89.35 & 88.89 & 88.63 & 86.94 \\
    \bottomrule
    \end{tabular}
    
\end{table}
}

\subsection{Ablations Studies}

\textbf{Impact of different components}. In~\Cref{ablation:component},  we investigate the effectiveness of the main components of our proposed AD-MT on the ACDC dataset with 5\% labeled data. The components studied include T1-only, T2-only, the RPA, and the CCM. The evaluation metrics used are Dice Score and 95HD, and we perform a category-wise examination for three classes: the right ventricle (RV), the myocardium (Myo), and the left ventricle (LV).
Recall that our AD-MT method iteratively updates two teacher models, named T1 and T2, where T1 is updated by the student model applying color augmentations and T2 is updated by the student model applying mix augmentations. 
It can be seen from the table that the T1-only obtains slightly better performance than the T2-only, indicating the superiority of the color or intensity based perturbation compared to the copy-paste augmentation. 
The third row shows the results when both T1 and T2 are involved, along with our proposed RPA module for alternate diverse updating. 
Although we only use the average prediction of two teacher models at the current stage, we observe a significant improvement in Dice Score for all three classes (RV, Myo, and LV), resulting in a mean improvement of more than 1\%.
It suggests that our RPA module can indeed generate diverse reasoning and consequently improve the segmentation performance.
Furthermore, as discussed in~\Cref{sec:intro}, two teacher models will inevitably come across conflicting supervision, and it is critical to address the conflicts.
As shown in the fourth row of the table, the complete AD-MT with all components, obtains the most accurate segmentation, with the lowest 95HD and the highest Dice Score.
Overall, the results demonstrate the importance of each component in our proposed AD-MT approach. Particularly, the RPA and CCM modules are effective in encouraging diverse reasoning and leveraging all involved models to improve the accuracy of the segmentation.
The results highlight the importance of leveraging diverse information from multiple sources and effectively combining them to improve segmentation accuracy.

\textbf{Impact of the threshold $\tau$}.
\Cref{tab:abl:threshold} shows the results of an ablation study on the pre-defined high-confidence threshold ($\tau$) for our AD-MT approach on the ACDC and LA in terms of the Dice Score.
For the ACDC, increasing the threshold from $0.7$ to $0.95$ leads to a gradual improvement in Dice Score, with the highest Dice Score of $88.75\%$ achieved at a threshold of $0.95$. Differently, for the LA dataset, increasing the threshold beyond $0.75$ leads to a drop in Dice Score. The highest Dice Score of $89.63\%$ is achieved at a threshold of $0.75$. It suggests that, on the 3D dataset, a higher threshold may filter out too many predictions, leading to a loss of information and a decrease in segmentation accuracy.
The default thresholds are $0.95$ and $0.75$ for the 2D and 3D datasets, respectively.

\begin{table}[t]
  \centering
  \begin{minipage}{0.57\linewidth}
    \centering
    \includegraphics[width=0.9\linewidth]{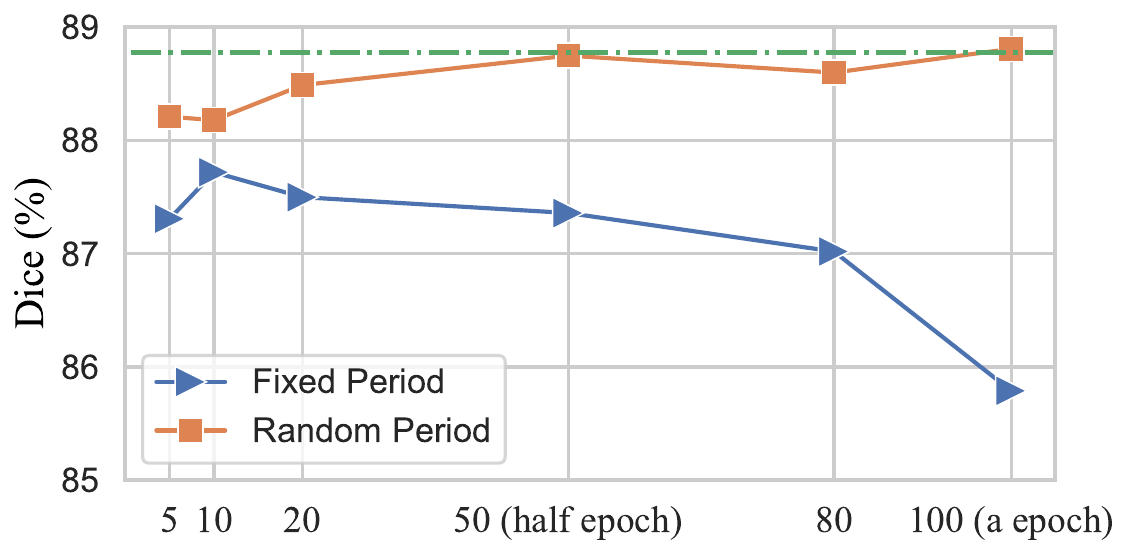}
    \captionof{figure}{Impact of different alternating periodic updating strategies in the RPA with varying values of $\mathcal{T}_{max}$ on the ACDC with 5\% labeled data. By default, we adopt the random switching periods and set $\mathcal{T}_{max}$ as the half-epoch iterations.}
    \label{fig:abl:period}
  \end{minipage}
  \hfill
  \begin{minipage}{0.4\linewidth}
    \centering
    \caption{Compare different ensembling strategies on the ACDC with 5\% labeled data, whenever conflicts occur. ``Drop" denotes dropping the conflicts completely.  ``Avg." and ``Ent." represent to use the mean and entropy-based ensembling of two teachers. }
    \label{tab:abl:ensembling}
    \begin{tabular}{c|cccc}
    \toprule
    Strategy & Drop & Avg. & Ent.  & CCM \\
    \midrule
    Dice (\%)  & 86.69  & 87.88 & 88.11 & \textbf{88.75} \\
    \bottomrule
    \end{tabular}
  \end{minipage}
\end{table}

\textbf{Impact of the switching periods $\mathcal{T}_{max}$}. \Cref{fig:abl:period} shows our examinations on different alternate updating periods in the RPA with two different strategies: the fixed and random periods. We can easily observe that the random strategies can consistently outperform the fixed ones. It is simply because the highly randomness can further enlarge the diversity between the two teacher models, resulting in better segmentation performance. In contrast, since both the student and teacher models are trained from scratch, large $\mathcal{T}_{max}$ in the fixed strategies, may enforce ensembling predictions among models with a big performance gap, which significantly reduces the ensembling effectiveness.

\textbf{Impact of different Ensembling strategies.} In~\Cref{tab:abl:ensembling}, we investigate different ensembling strategies to address the conflicts on the ACDC
with 5\% labeled data. We can clearly see that directly dropping all the conflicts leads to the lowest Dice Score of 86.69\%, indicating the significance of learning from the conflicts. Using the average and entropy-based ensembling strategies can clearly improve the segmentation performance, which is still limited at only involving the teacher models. 
As the training progresses,  the student model can effectively learn from both teachers and gradually become comparable to the teachers. Our CCM exploits such information and obtains the highest Dice of 88.75\%.

\begin{figure}[t]
  \centering
   \includegraphics[width=0.85\linewidth]{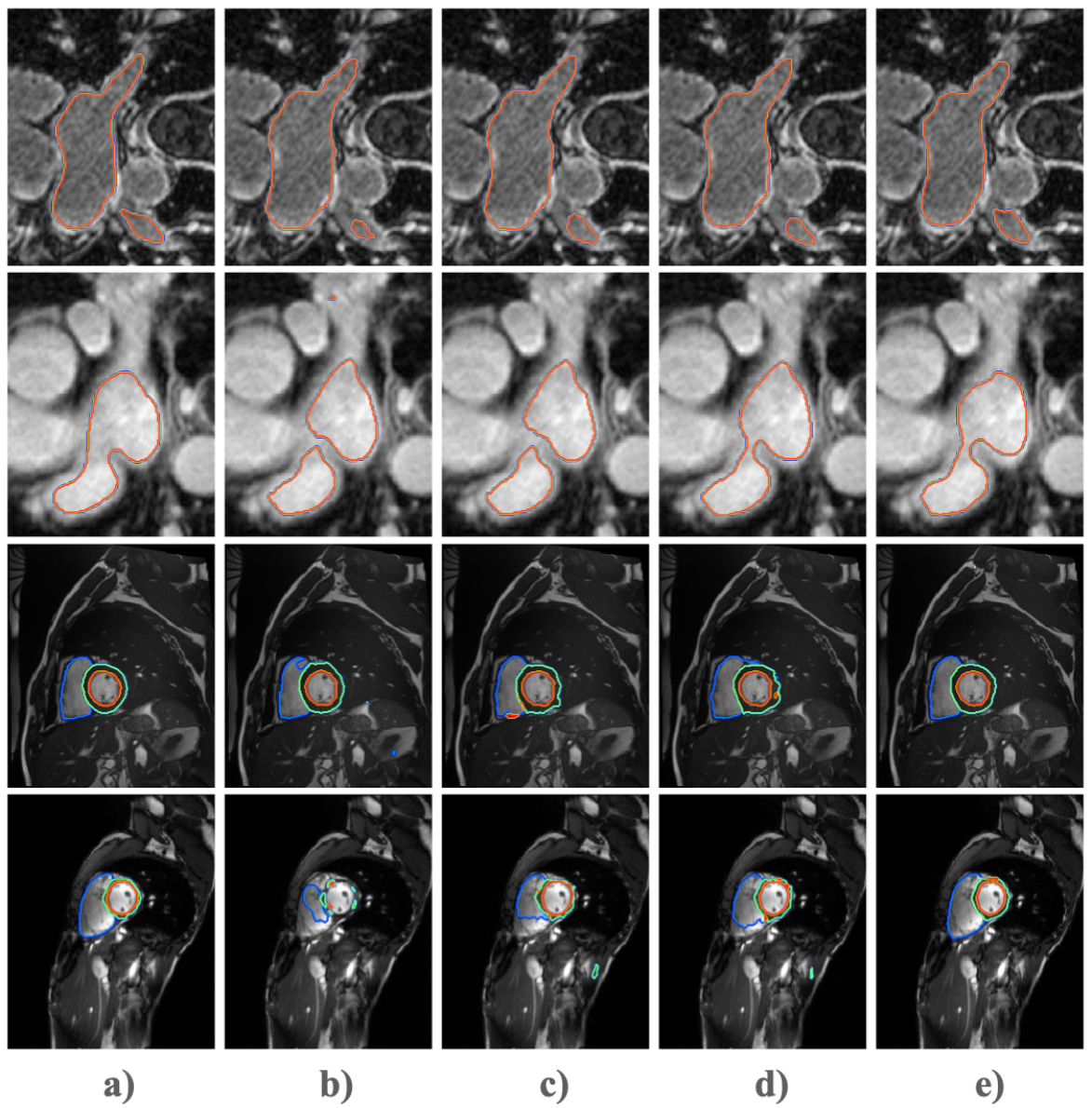}
   \caption{Qualitative results from the 3D LA (top 2 rows) and 2D ACDC (bottom 2 rows). a) the ground-truth, b) UA-MT, c) MC-Net, d) SS-Net, and e) AD-MT.}
   \label{fig:visual}
\end{figure}

\textbf{Visualization.}
\Cref{fig:visual} illustrates example segmentation results on the LA (top 2 rows)and ACDC (bottom 2 rows) in the semi-supervised setting of 5\%  labeled data. 
We clearly see that our approach produces segmentation results that are closer to the ground truth than the other SOTA methods. 
For example, only SS-Net and our AD-MT can recognize the connected segmented region, as shown in the second row. 
In contrast, SS-Net segments the wrong RV region in the third row on the multi-class ACDC dataset, while our ACDC does not.
Additionally, we observe that it is generally more challenging to segment three classes from a Sagittal plane (the fourth row) than from a Coronal plane (the third row) on the ACDC. The better segmentation results further highlight the importance of leveraging diverse information from multiple sources to improve segmentation accuracy, as our proposed AD-MT approach does by combining information from multiple modalities and teacher models. 
Overall, the results in \Cref{fig:visual} demonstrate the effectiveness of our proposed AD-MT approach in improving the accuracy of SSMIS.

\section{Conclusion}

In this paper, we propose an alternate diverse teaching approach in a teacher-student framework, which boosts SSMIS via two novel modules: the Random Periodic Alternate Updating Module and the Conflict-Combating Module.
With the RPA scheduling, two teacher models are momentum-updated periodically and randomly in an alternate manner to produce diverse supervision. The entropy-based CCM effectively leverages all involved models to encourage the student model to learn from the two teachers' consistent and conflicting predictions. Without introducing extra training parameters and constraints, AD-MT achieves the new SOTA performance on popular 2D and 3D SSMIS benchmarks.


\bibliographystyle{splncs04}
\bibliography{egbib}
\end{document}